%% file: main.tex
\documentclass[journal]{IEEEtran}
\usepackage{graphicx}
\usepackage{amsmath,amssymb}
\usepackage{booktabs}
\usepackage{multirow}
\usepackage{array}
\usepackage[table]{xcolor}
\usepackage{tikz}
\usetikzlibrary{positioning,arrows.meta,fit,backgrounds,calc}
\usepackage{pgfplots}
\pgfplotsset{compat=1.17}
\usepackage{microtype}
\usepackage{caption}
\usepackage[numbers,sort&compress]{natbib}
\definecolor{cvprblue}{rgb}{0.21,0.49,0.74}
\usepackage[breaklinks,colorlinks,allcolors=cvprblue]{hyperref}
\usepackage{balance}

\newcommand{\mrgb}{\mathrm{rgb}}
\newcommand{\mdep}{\mathrm{depth}}
\newcommand{\Trk}{\mathcal{T}}
\newcommand{\Mone}{\textsc{M1}}
\newcommand{\Mtwo}{\textsc{M2}}

\title{IIT Patna at ``\textit{ICPR 2026 TVRID Competition on Privacy-Preserving Person Re-Identification from Top-View RGB-Depth Camera}"}

\author{Abhinav Kumar, Swagatam Pati, Advait Deshmukh, Aditya Mehrotra, Bibek Das, Kumari Priya, \\ Utsav Kumar Nareti, Saba Akram, Chandranath Adak\\
{\footnotesize \{bibek\_2422cs27, utsav\_2221cs28\}@iitp.ac.in}
}

\begin{document}
\maketitle

%-------------------------------

\input{01abstract}
\input{02Introduction}
\input{03Method_1}
\input{04Method_02}

\input{05SetUp}

\input{06Dataset}

\input{07Result}
\input{08Conclusion}

\balance
%-------------------------------

{\footnotesize
\setlength{\bibsep}{0pt}
\bibliographystyle{IEEEtran}
\bibliography{refs}
}

\balance

\end{document}

%% file: 01abstract.tex
\begin{abstract}
Privacy-aware person re-identification (Re-ID) from top-view cameras replaces frontal appearance with anthropometric shape, while depth sensing further reduces color-based identity cues. The ICPR 2026 TVRID competition evaluates this setting using 86 identities captured by four synchronized Intel RealSense D455 cameras with RGB and Depth streams across flat, ascent, descent, and oblique viewpoints. Performance is measured using mAP and CMC@1 across within-camera in/out, ascent-descent, and flat-oblique scenarios. The competition had 147 registered participants and 33 final submissions. Three teams from IIT Patna participated: `abhinavkumar03' and `sjkdgfhl\_' ranked 5th and 7th among 31 ranked participants, and `vivek\_katuri' reached 77.67\% CMC@1 / 85.20\% mAP on the RGB track. This paper presents the methods of the two top-10 teams. Methodology 1 (abhinavkumar03) achieves 93.51\% mAP and 90.42\% CMC@1 on Depth, ranking 4th of 28, while Methodology 2 (sjkdgfhl\_) reaches 99.63\%/99.35\% on RGB and 86.26\%/80.34\% on Depth. Both substantially outperform the 57.19\% track-average Depth mAP, with Methodology 1 yielding a 7.3-point gain over Methodology 2. These results demonstrate the importance of depth representation and supervision for privacy-preserving Re-ID.
\end{abstract}


%% file: 02Introduction.tex
\section{Introduction}
\label{sec:intro}

Person re-identification (Re-ID) determines whether observations from different cameras belong to the same person and is widely used in surveillance, retail analytics, and crowd analysis~\cite{ye2022survey}. While most Re-ID methods rely on side-view RGB images and appearance cues such as clothing color and texture, overhead cameras provide a more privacy-aware alternative by reducing facial visibility, while depth sensors further reduce appearance information by capturing 3D body geometry without color~\cite{delecluse2025fg}. However, limited silhouettes, variable scale, short 2-3 second observations, and fewer color and texture cues make re-identification challenging. The ICPR~2026 TVRID competition~\cite{delecluse2026tvrid} evaluates this setting using synchronized RGB and depth data from four overhead cameras with different viewpoints. Three teams from IIT Patna participated, with two reaching the top 10. Methodology~1 ($\Mone$: abhinavkumar03) converts depth maps into JET colorized images and uses OSNet~\cite{zhou2019osnet}, while Methodology~2 ($\Mtwo$: sjkdgfhl\_) uses separate CNNs backbone for RGB and depth and mean-pools frame-level features to form a tracklet representation. Both methods use only the official TVRID training data and achieve strong performance on the competition benchmarks.

% The main approaches explored by the two teams are:
% (i) depth colorization with a lightweight OSNet backbone in \Mone; and
% (ii) modality-specific CNNs with tracklet-level feature aggregation in \Mtwo.

%% file: 03Method_1.tex
\input{fig_m1}
\input{fig_m2}

%% --------------

\textbf{Task formulation.} For each modality $m\in\{\mrgb,\mdep\}$, a tracklet $\Trk^m=\{x^m_t\}_{t=1}^{T}$ is the set of $T$ person crops from one camera pass, where $x^m_t$ is the crop at frame $t$. Using the $N{=}62$ labeled training identities, each method learns an embedding $v^m=\Psi^m(\Trk^m)\in\mathbb{R}^{d}$ with $\lVert v^m\rVert_2=1$ ($d{=}512$ for \Mone{}, $d{=}2048$ for \Mtwo). At test time, the identities are unseen. Each query tracklet $q$ is compared with every gallery tracklet $g$ using $D^m(q,g)=\lVert v^m_q-v^m_g\rVert_2$
and gallery tracklets are ranked by increasing distance.

%%---------------

\section{Methodology 1}
\label{sec:m1}

\textbf{Depth colorization:}
Since raw depth images are single-channel and lack the color and texture cues
used by ImageNet-pretrained CNNs, \Mone{} converts each depth frame into a
three-channel JET pseudo-color image using a fixed colormap $\mathcal{C}$ 
~\cite{eitel2015rgbd}:
\begin{equation}
\tilde{x}_t=\mathcal{C}(x^{\mdep}_t)
\in\mathbb{R}^{H\times W\times3}.
\label{eq:jet}
\end{equation}
This converts depth variations into color gradients and highlights structures
such as the head and shoulders without adding learnable parameters.

\textbf{OSNet backbone:}
The colorized image is resized to $256{\times}128$ and processed using
OSNet~\texttt{x1\_0}~\cite{zhou2019osnet}. Its omni-scale blocks capture
features at different spatial scales, helping handle changes in viewpoint,
elevation, and body scale in top-view images. Global average pooling and a
fully connected layer produce a 512-dimensional frame feature $f_t$. The
backbone has only 2.2M parameters.

\subsection{Training and Inference}
\label{sec:m1_train}

The network is trained to classify each frame into one of the $N{=}62$ training identities. For a mini-batch of $B$ frames, let $f_i\in\mathbb{R}^{512}$ be the embedding of frame $i$, $y_i$ its identity label, and $W\in\mathbb{R}^{N\times512}$ the weights of a linear classifier. The loss is label-smoothed cross-entropy:
\begin{equation}
\mathcal{L}_{\mathrm{LS}}=-\frac{1}{B}\sum_{i=1}^{B}\sum_{k=1}^{N}\tilde{y}_{i,k}\,\log\!\big[\mathrm{softmax}(Wf_i)_k\big],
\label{eq:ls}
\end{equation}
where the smoothed target is $\tilde{y}_{i,k}{=}(1-\varepsilon)\,\delta_{k,y_i}+\varepsilon/N$, $\delta_{k,y_i}$ equals 1 when $k{=}y_i$ and 0 otherwise, and $\varepsilon{=}0.1$. Training uses only the official TVRID training data, starting from ImageNet weights. We use Adam for 60 epochs with batch size 32, an initial learning rate of $3{\times}10^{-3}$ and weight decay $5{\times}10^{-4}$, and reduce the learning rate by $10\times$ at epochs 30 and 50. For augmentation, each image is flipped horizontally with probability 0.5. With probability 0.5 it is also enlarged to $288{\times}144$ and randomly cropped back to $256{\times}128$; otherwise it is resized directly to $256{\times}128$.

During inference, the classifier is removed and each frame feature is
$\ell_2$-normalized: 
% \begin{equation}
$v_t^{\mdep}{=}{f_t}/{\|f_t\|_2}$.
% \end{equation}
The resulting frame embeddings are used to represent the tracklets, and
query-gallery pairs are ranked using $D^m(q,g)$.

%% file: fig_m1.tex
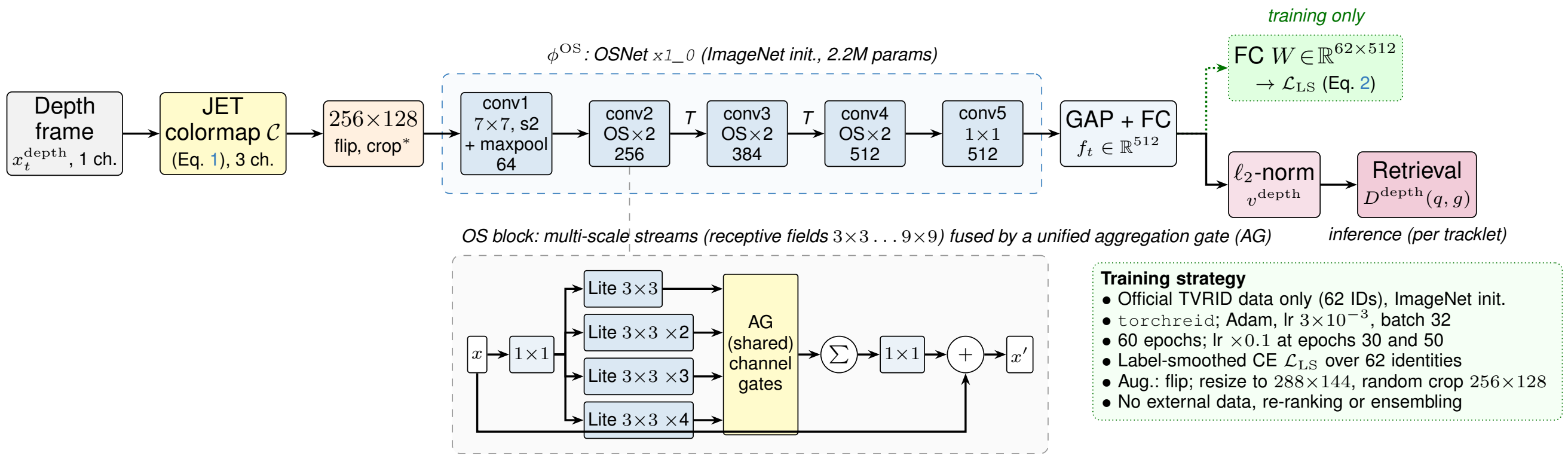
\begin{figure*}[!t]
\centering
\resizebox{0.95\linewidth}{!}{%
\begin{tikzpicture}[
  font=\sffamily\small, node distance=2.5mm and 5mm,
  blk/.style={draw, rounded corners=2pt, minimum height=9mm, inner sep=2pt, align=center, fill=#1},
  blk/.default=white,
  st/.style={draw, rounded corners=1.5pt, minimum height=9mm, minimum width=11mm, inner sep=1.5pt, align=center, fill=cvprblue!18, font=\sffamily\scriptsize},
  tr/.style={draw, rounded corners=1.5pt, minimum height=9mm, inner sep=1.5pt, align=center, fill=cvprblue!6, font=\sffamily\scriptsize},
  sm/.style={draw, rounded corners=1pt, minimum height=5mm, inner sep=1.5pt, align=center, fill=#1, font=\sffamily\scriptsize},
  arr/.style={-{Stealth[length=1.6mm]}, thick},
  trn/.style={-{Stealth[length=1.6mm]}, thick, densely dotted, draw=green!45!black},
  lbl/.style={font=\sffamily\scriptsize\itshape}]
% ---------------- main row ----------------
\node[blk=gray!10] (in) {Depth\\frame\\{\scriptsize $x^{\mdep}_t$, 1 ch.}};
\node[blk=yellow!25, right=of in] (jet) {JET\\colormap $\mathcal{C}$\\{\scriptsize (Eq.~\ref{eq:jet}), 3 ch.}};
\node[blk=orange!12, right=of jet] (aug) {$256{\times}128$\\{\scriptsize flip, crop$^{\ast}$}};
\node[st, right=5mm of aug] (c1) {conv1\\$7{\times}7$, s2\\+ maxpool\\64};
\node[st, right=of c1] (c2) {conv2\\OS$\times$2\\256};
\node[st, right=of c2] (c3) {conv3\\OS$\times$2\\384};
\node[st, right=of c3] (c4) {conv4\\OS$\times$2\\512};
\node[st, right=of c4] (c5) {conv5\\$1{\times}1$\\512};
\node[blk=cvprblue!8, right=of c5] (gap) {GAP + FC\\{\scriptsize $f_t\in\mathbb{R}^{512}$}};
\node[blk=purple!12, right=7mm of gap, yshift=-7mm] (l2) {$\ell_2$-norm\\{\scriptsize $v^{\mdep}$}};
\node[blk=purple!20, right=of l2] (ret) {Retrieval\\{\scriptsize $D^{\mdep}(q,g)$}};
\node[blk=green!12, densely dotted, draw=green!45!black, right=7mm of gap, yshift=9mm] (head) {FC $W\!\in\!\mathbb{R}^{62\times512}$\\{\scriptsize $\to\mathcal{L}_{\mathrm{LS}}$ (Eq.~\ref{eq:ls})}};
\begin{scope}[on background layer]
\node[draw=cvprblue, dashed, rounded corners=3pt, fill=cvprblue!3, inner sep=2.5mm, fit=(c1)(c5)] (osn) {};
\end{scope}
\node[lbl, above=0mm of osn.north, anchor=south] {$\phi^{\mathrm{OS}}$: OSNet \texttt{x1\_0} (ImageNet init., 2.2M params)};
\draw[arr] (in)--(jet); \draw[arr] (jet)--(aug); \draw[arr] (aug)--(c1);
\draw[arr] (c1)--(c2); \draw[arr] (c2)-- node[lbl, above]{T} (c3); \draw[arr] (c3)-- node[lbl, above]{T} (c4); \draw[arr] (c4)--(c5); \draw[arr] (c5)--(gap);
\draw[trn] (gap.east) -- ++(4mm,0) |- (head.west);
\draw[arr] (gap.east) -- ++(4mm,0) |- (l2.west);
\draw[arr] (l2)--(ret);
\node[lbl, text=green!45!black, above=0mm of head] {training only};
\node[lbl, below=0mm of ret] {inference (per tracklet)};
% ---------------- OS block inset ----------------
\node[sm=white, below=22mm of c1, xshift=-4mm] (bi) {$x$};
\node[sm=cvprblue!10, right=3mm of bi] (p1) {$1{\times}1$};
\node[sm=cvprblue!18, right=4mm of p1, yshift=9mm]  (s1) {Lite $3{\times}3$};
\node[sm=cvprblue!18, right=4mm of p1, yshift=3mm]  (s2) {Lite $3{\times}3$ $\times$2};
\node[sm=cvprblue!18, right=4mm of p1, yshift=-3mm] (s3) {Lite $3{\times}3$ $\times$3};
\node[sm=cvprblue!18, right=4mm of p1, yshift=-9mm] (s4) {Lite $3{\times}3$ $\times$4};
\node[sm=yellow!25, minimum height=22mm, anchor=west] (ag) at ($(s4.east |- p1)+(4mm,0)$) {AG\\(shared)\\{\scriptsize channel}\\{\scriptsize gates}};
\node[sm=white, right=3mm of ag, circle, inner sep=0.5pt] (sum) {$\sum$};
\node[sm=cvprblue!10, right=3mm of sum] (p2) {$1{\times}1$};
\node[sm=white, right=3mm of p2, circle, inner sep=0.5pt] (add) {$+$};
\node[sm=white, right=3mm of add] (bo) {$x'$};
\foreach \s in {s1,s2,s3,s4}{\draw[arr] (p1.east) -- ++(1.5mm,0) |- (\s.west); \draw[arr] (\s.east) -- (\s.east -| ag.west);}
\draw[arr] (bi)--(p1); \draw[arr] (ag)--(sum); \draw[arr] (sum)--(p2); \draw[arr] (p2)--(add); \draw[arr] (add)--(bo);
\draw[arr] (bi.south) -- ++(0,-8mm) -| (add.south);
\begin{scope}[on background layer]
\node[draw=black!40, dashed, rounded corners=3pt, fill=black!2, inner sep=2mm, fit=(bi)(s1)(s4)(bo)] (inset) {};
\end{scope}
\node[lbl, above=0mm of inset.north west, anchor=south west] {OS block: multi-scale streams (receptive fields $3{\times}3\ldots9{\times}9$) fused by a unified aggregation gate (AG)};
\draw[black!40, dashed] (c2.south) -- (inset.north -| c2.south);
% ---------------- training strategy box ----------------
\node[draw=green!45!black, densely dotted, rounded corners=3pt, fill=green!4, align=left, font=\sffamily\scriptsize, anchor=north west, text width=62mm] (tsbox) at ($(inset.north east)+(6mm,-1mm)$)
{\textbf{Training strategy}\\
$\bullet$ Official TVRID data only (62 IDs), ImageNet init.\\
$\bullet$ \texttt{torchreid}; Adam, lr $3{\times}10^{-3}$, batch 32\\
$\bullet$ 60 epochs; lr $\times0.1$ at epochs 30 and 50\\
$\bullet$ Label-smoothed CE $\mathcal{L}_{\mathrm{LS}}$ over 62 identities\\
$\bullet$ Aug.: flip; resize to $288{\times}144$, random crop $256{\times}128$\\
$\bullet$ No external data, re-ranking or ensembling};
\end{tikzpicture}}
\caption{\textbf{Methodology~1} (abhinavkumar03): Depth frames are JET-colorized and encoded by OSNet \cite{zhou2019osnet}}
\label{fig:m1}
\end{figure*}

%% file: fig_m2.tex
\begin{figure*}[!t]
\centering
\resizebox{0.95\linewidth}{!}{%
\begin{tikzpicture}[
  font=\sffamily\small, node distance=2.5mm and 3.5mm,
  blk/.style={draw, rounded corners=2pt, minimum height=9mm, inner sep=2pt, align=center, fill=#1},
  blk/.default=white,
  st/.style={draw, rounded corners=1.5pt, minimum height=8mm, minimum width=10mm, inner sep=1.5pt, align=center, fill=#1, font=\sffamily\scriptsize},
  sh/.style={minimum height=26mm},
  arr/.style={-{Stealth[length=1.6mm]}, thick},
  trn/.style={-{Stealth[length=1.6mm]}, thick, densely dotted, draw=green!45!black},
  lbl/.style={font=\sffamily\scriptsize\itshape}]
% ---------------- inputs ----------------
\node[blk=gray!10] (inR) at (0, 8mm) {RGB\\tracklet\\{\scriptsize $\Trk^{\mrgb}$}};
\node[blk=gray!10] (inD) at (0,-8mm) {Depth\\tracklet\\{\scriptsize $\Trk^{\mdep}$}};
\node[blk=orange!12, sh, anchor=west] (aug) at ($(inR.east)!0.5!(inD.east)+(5mm,0)$) {$256{\times}128$\\[2pt]{\scriptsize flip $p{=}.5$}\\{\scriptsize rot.\ $\pm10^\circ$}\\{\scriptsize jitter}\\{\scriptsize erasing $p{=}.5$}$^{\ast}$};
% ---------------- RGB backbone stages ----------------
\node[st=cvprblue!22, anchor=west] (r0) at ($(aug.east)+(6mm,8mm)$) {deep\\stem\\64};
\node[st=cvprblue!22, right=of r0] (r1) {S1\\SA$\times$3\\256};
\node[st=cvprblue!22, right=of r1] (r2) {S2\\SA$\times$4\\512};
\node[st=cvprblue!22, right=of r2] (r3) {S3\\SA$\times$6\\1024};
\node[st=cvprblue!22, right=of r3] (r4) {S4\\SA$\times$3\\2048};
% ---------------- Depth backbone stages ----------------
\node[st=cvprblue!12] (d0) at (r0 |- inD) {$7{\times}7$\\stem\\64};
\node[st=cvprblue!12] (d1) at (r1 |- inD) {S1\\B$\times$3\\256};
\node[st=cvprblue!12] (d2) at (r2 |- inD) {S2\\B$\times$4\\512};
\node[st=cvprblue!12] (d3) at (r3 |- inD) {S3\\B$\times$6\\1024};
\node[st=cvprblue!12] (d4) at (r4 |- inD) {S4\\B$\times$3\\2048};
\begin{scope}[on background layer]
\node[draw=cvprblue, dashed, rounded corners=3pt, fill=cvprblue!3, inner sep=2mm, fit=(r0)(r4)] (bbR) {};
\node[draw=cvprblue, dashed, rounded corners=3pt, fill=cvprblue!3, inner sep=2mm, fit=(d0)(d4)] (bbD) {};
\end{scope}
\node[lbl, above=0mm of bbR.north, anchor=south] {$\phi^{\mrgb}$: ResNeSt-50d (split-attention, radix 2), 27M};
\node[lbl, below=0mm of bbD.south, anchor=north] {$\phi^{\mdep}$: ResNet-50 (bottleneck blocks), 26M};
% ---------------- shared stages ----------------
\node[blk=cvprblue!8, sh, anchor=west] (gap) at ($(bbR.east)!0.5!(bbD.east)+(6mm,0)$) {GAP\\[2pt]{\scriptsize $f_t\in\mathbb{R}^{2048}$}};
\node[blk=purple!10, sh, right=8mm of gap] (agg) {Mean pool\\over $t$\\+ $\ell_2$\\[2pt]{\scriptsize $v^{m}$ (Eq.~\ref{eq:agg})}};
\node[blk=purple!18, sh, right=of agg] (dist) {Retrieval\\[2pt]{\scriptsize $D^{m}(q,g)$}};
\node[blk=red!12, sh, right=7mm of dist] (fuse) {Score\\fusion\\[2pt]{\scriptsize $D^{\mathrm{fused}}$}\\{\scriptsize (Eq.~\ref{eq:fuse})}};
\node[blk=green!12, densely dotted, draw=green!45!black, above=5mm of gap, xshift=6mm] (head) {BN1d $\to W\!\in\!\mathbb{R}^{62\times2048}$ $\to\mathcal{L}_{\mathrm{ID}}$ (Eq.~\ref{eq:id})};
\node[lbl, text=green!45!black, above=0mm of head] {training only (per frame)};
% ---------------- arrows ----------------
\draw[arr] (inR.east) -- (inR.east -| aug.west);
\draw[arr] (inD.east) -- (inD.east -| aug.west);
\draw[arr] (aug.east |- r0) -- (r0.west);
\draw[arr] (aug.east |- d0) -- (d0.west);
\foreach \a/\b in {r0/r1,r1/r2,r2/r3,r3/r4,d0/d1,d1/d2,d2/d3,d3/d4}{\draw[arr] (\a)--(\b);}
\draw[arr] (r4.east) -- (r4.east -| gap.west);
\draw[arr] (d4.east) -- (d4.east -| gap.west);
\draw[trn] (gap.north) -- (gap.north |- head.south);
\draw[arr] (gap) -- (agg);
\draw[arr] (agg) -- (dist);
\draw[arr] ([yshift=5mm]dist.east) -- node[lbl, above]{rgb} ([yshift=5mm]fuse.west);
\draw[arr] ([yshift=-5mm]dist.east) -- node[lbl, below]{dep} ([yshift=-5mm]fuse.west);
\node[lbl, below=0mm of $(gap.south)!0.5!(dist.south)$, text=black!70] {shared form, separate weights per stream $m$};
\node[lbl, below=0mm of agg.south, yshift=-3.5mm] {inference (per tracklet)};
% ---------------- training strategy box ----------------
\node[draw=green!45!black, densely dotted, rounded corners=3pt, fill=green!4, align=left, font=\sffamily\scriptsize, anchor=north west, text width=88mm] (tsbox) at ($(bbD.south west)+(0,-6mm)$)
{\textbf{Training strategy:} official TVRID data only (62 IDs); ImageNet init.; one model per modality; frame-level CE $\mathcal{L}_{\mathrm{ID}}$ with BN neck; Adam, lr $3.5{\times}10^{-4}$, mixed precision; aug.\ flip, rotation, color jitter, random erasing.};
\end{tikzpicture}}
\caption{\textbf{Methodology~2} (sjkdgfhl\_): Modality-specific CNN backbone encodes frames, followed by mean pooling for tracklet retrieval \cite{he2016resnet}}
\label{fig:m2}
\end{figure*}
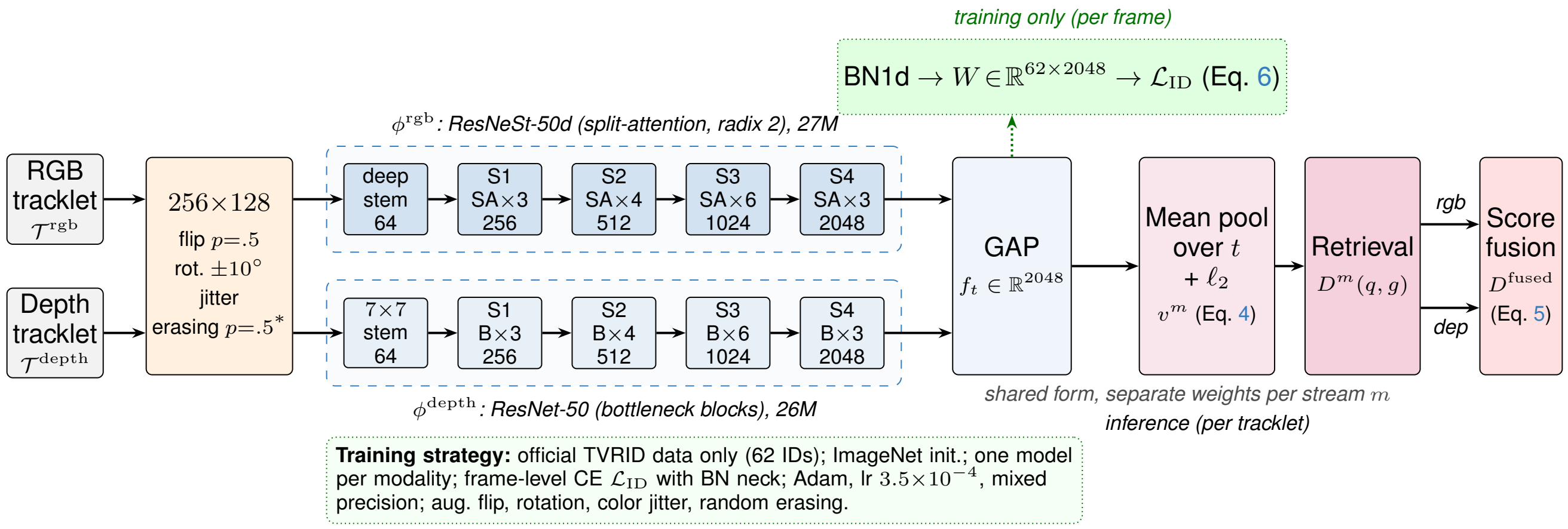

%% file: 04Method_02.tex
\section{Methodology 2}
\label{sec:m2}

\textbf{Modality-specific backbones:} For each modality $m\in\{\mrgb,\mdep\}$, every frame $x^m_t$ of a tracklet is resized to $256{\times}128$ and encoded independently by its own backbone $\phi^m$, a convolutional network with its ImageNet classification head removed. We use ResNeSt-50d~\cite{zhang2022resnest} for RGB ($\phi^{\mrgb}$) and ResNet-50~\cite{he2016resnet} for depth ($\phi^{\mdep}$). Both start from ImageNet weights, and depth frames are read as grayscale images replicated to three channels so that they match the RGB input format. The backbone outputs a feature map $\phi^m(x^m_t)\in\mathbb{R}^{2048\times h\times w}$, where $h\times w$ is its spatial size. Global average pooling (GAP), which averages this map over all $h\times w$ positions, gives a 2048-dimensional frame feature:
\begin{equation}
f_t=\mathrm{GAP}\big(\phi^m(x^m_t)\big)\in\mathbb{R}^{2048}
\label{eq:m2_frame}
\end{equation}

\textbf{Tracklet aggregation:} At test time the classifier $W$ is removed. A tracklet $\Trk^m$ contains $T$ frames, which are short and unordered, so instead of modeling time we average the $T$ frame features $f_t$ from Eq.~\ref{eq:m2_frame} and scale the result to unit length:
\begin{equation}
\bar f=\frac{1}{T}\sum_{t=1}^{T}f_t,
\qquad
v^m=\frac{\bar f}{\lVert\bar f\rVert_2},
\label{eq:agg}
\end{equation}
where $\lVert\cdot\rVert_2$ is the Euclidean ($\ell_2$) norm. For depth, the BN layer of the training head is kept, so each $f_t$ is batch-normalized before averaging. Averaging adds no parameters, does not depend on frame order or on $T$, and smooths out frame-level noise such as motion blur, partial views and detector errors. The tracklet embedding $v^m$ is then compared across query and gallery tracklets with $D^m(q,g)$.

\textbf{Score-level fusion:}
When both RGB and depth are available, their query-gallery distances are averaged:
\begin{equation}
D^{\mathrm{fused}}
=
\frac{1}{2}
\left(
D^{\mrgb}+D^{\mdep}
\right)
\label{eq:fuse}
\end{equation}
This provides RGB+Depth matching using both modalities for the query and gallery.

\subsection{Training Strategy}
\label{sec:m2_train}
Each modality branch is trained independently to classify frames into the $N{=}62$ training identities. For a mini-batch $\mathcal{B}$ of frame--label pairs $(x^m_t,y)$, the frame feature $f_t$ (Eq.~\ref{eq:m2_frame}) passes through a BN layer and a linear classifier $W\in\mathbb{R}^{N\times2048}$, trained with cross-entropy:
\begin{equation}
\mathcal{L}_{\mathrm{ID}}=-\frac{1}{|\mathcal{B}|}\sum_{(x^m_t,\,y)\in\mathcal{B}}\log\Big[\mathrm{softmax}\big(W\,\mathrm{BN}(f_t)\big)_{y}\Big]
\label{eq:id}
\end{equation}
We use only TVRID training data with ImageNet initialization and train with Adam (constant learning rate $3.5{\times}10^{-4}$, batch size 64). Augmentation includes horizontal flipping, rotation ($\pm10^\circ$), color jitter and random erasing~\cite{zhong2020erasing}. During inference, the BatchNorm layer and classifier are removed. Frame features are mean-pooled using Eq.~\ref{eq:agg} to obtain the $\ell_2$-normalized tracklet embedding. When both RGB and depth are available, their distances are fused using Eq.~\ref{eq:fuse}.

%% file: 05SetUp.tex
%-------------------------------
\section{Experimental Setup}
\label{sec:setup}

\textbf{Common protocol:} Both methods use ImageNet initialization and train only on the 62 official TVRID training identities. We evaluate final rankings on the blind \href{https://www.codabench.org/competitions/12315/#/results-tab}{Codabench server} using CMC@1 and mAP~\cite{delecluse2026tvrid}.

\textbf{Implementation:} Table~\ref{tab:impl} summarizes the two configurations. \Mone{} \& \Mtwo{} are implemented and trained on the Aryabhatta Paramrudra Server.
 
\begin{table}[!t]
\centering
\scriptsize 
\setlength{\tabcolsep}{2.5pt}
\renewcommand{\arraystretch}{0.95}
\begin{tabular}{@{}l>{\raggedright\arraybackslash}p{0.36\linewidth}>{\raggedright\arraybackslash}p{0.43\linewidth}@{}}
\toprule
 & \Mone{} (\texttt{abhinavkumar03}) & \Mtwo{} (\texttt{sjkdgfhl\_}) \\
\midrule
Input & Depth $\rightarrow$ JET pseudo-RGB & RGB / raw depth \\
Backbone & OSNet \texttt{x1\_0} (2.2M) & ResNeSt-50d (27M) / ResNet-50 (26M) \\
Embedding & 512-d, $\ell_2$ & 2048-d, mean-pooled, $\ell_2$ \\
Loss & $\mathcal{L}_{\mathrm{LS}}$ & $\mathcal{L}_{\mathrm{ID}}$ \\
Optimizer & Adam, lr $3{\times}10^{-3}$ & Adam, lr $3.5{\times}10^{-4}$ \\
Schedule & 60 ep., batch 32; $\times0.1$ at 30, 50 & -- \\
Augmentation & Flip; crop from $288{\times}144$ & Flip, rotation $\pm10^\circ$, jitter, erasing \\
\bottomrule
\end{tabular}
\caption{Implementation details of the two methods.}
\label{tab:impl}
\end{table}

%% file: 06Dataset.tex
\begin{table}[!t]
\centering
\footnotesize
\setlength{\tabcolsep}{4pt}
\begin{tabular}{@{}llcccc@{}}
\toprule
Track & Metric & Average & Median & \Mone & \Mtwo \\
\midrule
\multirow{2}{*}{RGB}   & mAP   & 89.66 & 93.35 & \textbf{99.88} & 99.63 \\
                       & CMC@1 & 85.73 & 91.92 & \textbf{99.84} & 99.35 \\
\midrule
\multirow{2}{*}{Depth} & mAP   & 57.19 & 49.82 & \textbf{93.51} & 86.26 \\
                       & CMC@1 & 47.10 & 30.55 & \textbf{90.42} & 80.34 \\
\bottomrule
\end{tabular}
\caption{Our entries against the official aggregate statistics~\cite{delecluse2026tvrid}}
\label{tab:agg}
\end{table}

%% ----------------------------------  

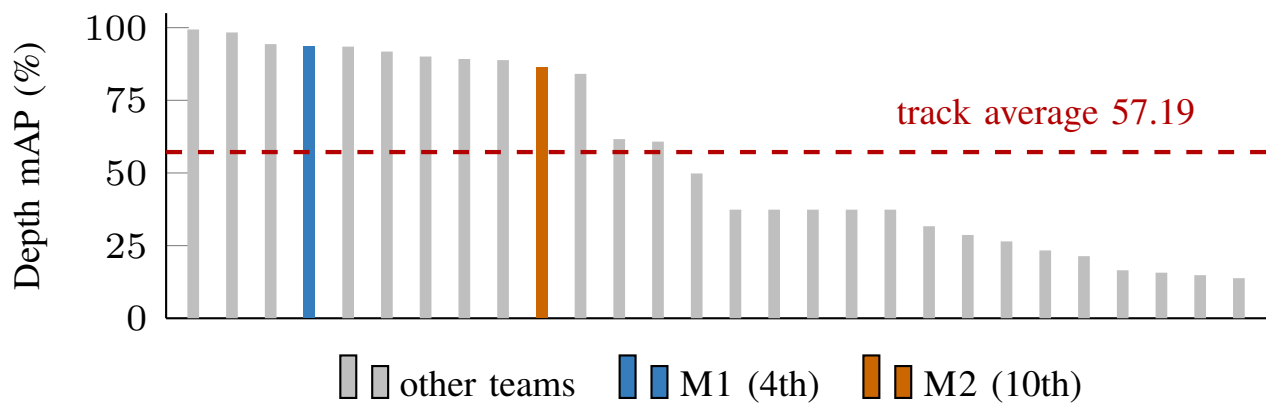
\begin{figure}[!t]
\centering
\begin{tikzpicture}
\begin{axis}[
  width=\linewidth, height=3.6cm,
  ybar, bar width=2.2pt, bar shift=0pt, ymin=0, ymax=105,
  xmin=0.3, xmax=28.7,
  xtick=\empty, ytick={0,25,50,75,100},
  ylabel={Depth mAP (\%)}, ylabel style={font=\scriptsize},
  yticklabel style={font=\scriptsize},
  axis lines*=left, enlarge x limits=false,
  legend style={font=\scriptsize, draw=none, at={(0.5,-0.08)}, anchor=north, fill=none, legend columns=3, /tikz/every even column/.append style={column sep=6pt}},
  legend cell align=left]
\addplot[fill=black!25, draw=none] coordinates {
 (1,99.37)(2,98.34)(3,94.33)(5,93.47)(6,91.77)(7,90.06)(8,89.22)(9,88.83)
 (11,84.10)(12,61.64)(13,60.78)(14,49.83)(15,37.37)(16,37.37)(17,37.37)(18,37.37)
 (19,37.37)(20,31.67)(21,28.67)(22,26.45)(23,23.35)(24,21.37)(25,16.51)(26,15.69)
 (27,14.83)(28,13.79)};
\addlegendentry{other teams}
\addplot[fill=cvprblue, draw=none] coordinates {(4,93.51)};
\addlegendentry{\Mone{} (4th)}
\addplot[fill=orange!80!black, draw=none] coordinates {(10,86.26)};
\addlegendentry{\Mtwo{} (10th)}
\draw[red!70!black, dashed, thick] (axis cs:0.3,57.19) -- (axis cs:28.7,57.19)
  node[pos=0.8, above, font=\scriptsize, text=red!70!black] {track average 57.19};
\end{axis}
\end{tikzpicture}
\caption{Official Depth-track mAP for 28 entries~\cite{delecluse2026tvrid}, sorted in decreasing order. Our entries are in the top tier, above the track average}
\label{fig:depthbar}
\end{figure}

\subsection{Dataset Analysis:}
\label{sec:data}

 TVRID~\cite{delecluse2026tvrid} contains 86 identities recorded by four synchronized overhead Intel RealSense D455 cameras at 15 FPS (640$\times$480), covering flat, ascent, descent, and oblique views. Each passage is observed twice (\textsc{in}/\textsc{out}), giving eight observation sets, with aligned RGB and depth data and depth-based person crops. The dataset uses 62 identities for training, with disjoint test identities. The main challenges are short 2-3 sec  tracklets, viewpoint and scale changes, and limited color and texture in depth images. Evaluation considers within- camera \textsc{in}$\leftrightarrow$\textsc{out}, ascent$\leftrightarrow$descent, and flat$\leftrightarrow$oblique scenarios using CMC@1 and mAP. The competition had 147 registered participants, 732 development submissions, and 33 final submissions, with average RGB mAP/CMC@1 of 89.66\%/85.73\% compared with 57.19\%/47.10\% for depth, highlighting the greater difficulty of depth-based Re-ID.

%% file: 07Result.tex
%-------------------------------
\subsection{Result Analysis:}
\label{sec:results}

\Mone{} and \Mtwo{} achieve strong performance across both tracks, obtaining 99.88\% and 99.63\% RGB mAP with 99.84\% and 99.35\% CMC@1, respectively, and 93.51\% and 86.26\% Depth mAP with 90.42\% and 80.34\% CMC@1. Both exceed the official Depth-track averages of 57.19\% mAP and 47.10\% CMC@1~\cite{delecluse2026tvrid}. On Depth, \Mone{} outperforms \Mtwo{} by 7.25 mAP and 10.08 CMC@1 points despite using a much smaller backbone (2.2M vs.\ 26M parameters), securing 4th rank among 28 submissions. This highlights the effectiveness of JET-based depth colorization with lightweight multi-scale OSNet. Both methods use only official TVRID training data and standard CNN-based training; while \Mone{} provides a lightweight depth solution, \Mtwo{} employs modality-specific backbones and tracklet-level feature aggregation. Both process RGB and Depth independently without requiring cross-modal alignment during training. However, the lack of controlled ablations and differences in training settings limit the attribution of performance gains to individual components.

% \Mone{} and \Mtwo{} achieve strong performance on both RGB and Depth tracks. On RGB, they obtain 99.88\% and 99.63\% mAP, respectively, with CMC@1 scores of 99.84\% and 99.35\%. On Depth, \Mone{} achieves 93.51\% mAP and 90.42\% CMC@1, while \Mtwo{} achieves 86.26\% mAP and 80.34\% CMC@1. Both depth results are well above the official depth-track average of 57.19\% mAP and 47.10\% CMC@1~\cite{delecluse2026tvrid}.
% %
% The depth results show a clear difference between the two approaches. \Mone{} outperforms \Mtwo{} by 7.25 mAP and 10.08 CMC@1 while using a much smaller backbone (2.2M vs.\ 26M parameters). This suggests that the JET colorization of depth, combined with the lightweight multi-scale OSNet, is effective for depth-based Re-ID. \Mone{} ranks 4th among 28 depth submissions, showing that a lightweight CNN can perform competitively without external data or complex training strategies.
% %
% Overall, both methods use only the official TVRID training data and standard CNN-based training. \Mone{} provides a lightweight depth-based solution, while \Mtwo{} benefits from modality-specific backbones and tracklet-level feature aggregation. Both approaches process RGB and depth independently, supporting privacy-aware Re-ID without requiring RGB-depth alignment during training. However, controlled ablations were not possible during the competition, and differences in training settings prevent a direct attribution of performance gains to individual components.

%% file: 08Conclusion.tex
%-------------------------------
\section{Conclusion}
\label{sec:conclusion}

We presented two IIT Patna entries in the ICPR~2026 TVRID competition using only the official training data. \Mone{} combines JET depth colorization with a lightweight 2.2M-parameter OSNet, while \Mtwo{} uses modality-specific backbones (ResNeSt-50d \& ResNet-50) with tracklet aggregation. They achieve up to 99.88\% mAP on RGB and 93.51\% on Depth, with \Mone{} ranking 5th overall and 4th on the Depth track, while \Mtwo{} ranks 7th overall. The results show that effective depth representation and simple feature aggregation can provide strong Re-ID performance without large models or external data.

\textbf{Acknowledgment:} 
The organizers of the ICPR 2026 TVRID Competition and the Aryabhatta Supercomputing Centre (ASC), IIT Patna, established under NSM, Govt. of India, for providing computational resources, are gratefully acknowledged.

%% file: refs.bib
@ARTICLE{ye2022survey,
  author={Ye, Mang and others},
  journal={IEEE TPAMI}, 
  title={{Deep Learning for Person Re-Identification: A Survey and Outlook}}, 
  year={2022},
  volume={44},
  number={6},
  pages={2872-2893}}

@INPROCEEDINGS{delecluse2025fg,
  author={Delécluse, Raphaël and others},
  booktitle={IEEE FG}, 
  title={{Privacy-Preserving Person Re-Identification from Temporal Sequences with Transformer and Hungarian Optimization}}, 
  year={2025},
  pages={1-10}}

@INPROCEEDINGS{eitel2015rgbd,
  author={Eitel, Andreas and others},
  booktitle={IEEE IROS}, 
  title={{Multimodal Deep Learning for Robust RGB-D Object Recognition}}, 
  year={2015},
  pages={681-687}}

@InProceedings{he2016resnet,
author = {He, Kaiming and others},
title = {{Deep Residual Learning for Image Recognition}},
booktitle = {CVPR},
year = {2016}
}

@INPROCEEDINGS{zhang2022resnest,
  author={Zhang, Hang and others},
  booktitle={IEEE CVPRW}, 
  title={{ResNeSt: Split-Attention Networks}}, 
  year={2022},
  pages={2735-2745}}

@InProceedings{delecluse2026tvrid,
author="Del{\'e}cluse, Rapha{\"e}l and others",
title="ICPR 2026 Competition on Privacy-Preserving Person Re-identification from Top-View RGB-Depth Camera (TVRID)",
booktitle="Pattern Recognition",
year="2027",
pages="229--242",
isbn="978-3-032-31936-4"
}

@inproceedings{zhou2019osnet,
  title={Omni-Scale Feature Learning for Person Re-Identification},
  author={Zhou, Kaiyang and Yang, Yongxin and Cavallaro, Andrea and Xiang, Tao},
  booktitle={ICCV},
  year={2019}
}

@inproceedings{zhong2020erasing,
  title={Random Erasing Data Augmentation},
  author={Zhong, Zhun and Zheng, Liang and Kang, Guoliang and Li, Shaozi and Yang, Yi},
  booktitle={AAAI},
  year={2020}
}
